\pdfoutput=1
\documentclass[10pt, logo]{template/nvidiatechreport}

\usepackage{natbib}
\usepackage{hyperref}
\usepackage{xspace}
\usepackage{amsmath}
\usepackage{amssymb}
\usepackage{graphicx}
\usepackage[dvipsnames]{xcolor}
\usepackage{listings}
\usepackage{multirow}
\usepackage{booktabs}
\usepackage{fontawesome5}
\usepackage{float}
\usepackage{wrapfig}

\definecolor{githubgray}{RGB}{240,240,240}      
\definecolor{githubblue}{RGB}{0,92,197}
\definecolor{githubgreen}{RGB}{22,163,74}
\definecolor{githubred}{RGB}{207,34,46}
\definecolor{githubpurple}{RGB}{136,46,224}
\lstdefinestyle{github}{
    language=Python,
    basicstyle=\ttfamily\small,             
    keywordstyle=\color{githubpurple},           
    commentstyle=\color{githubgreen}\itshape,    
    stringstyle=\color{githubred},               
    numberstyle=\tiny\color{gray!60},            
    numbers=left,
    stepnumber=1,
    numbersep=8pt,                              
    showstringspaces=false,
    breaklines=true,
    frame=none,                                  
    backgroundcolor=\color{githubgray},          
    xleftmargin=0pt,                            
    framexleftmargin=0pt,
    rulecolor=\color{gray!30},
    tabsize=2,                                  
    captionpos=b,
    basewidth=0.5em,                            
    columns=fixed
}

\definecolor{darkblue}{rgb}{0, 0, 0.3}
\hypersetup{colorlinks=true, citecolor=darkblue, linkcolor=darkblue, urlcolor=darkblue}

\newcommand{\githubrepo}{\href{https://github.com/NVIDIA/kvpress}{{\faGithub} NVIDIA/kvpress}\xspace}

\newcommand{\hflink}[2]{%
  \href{#1}{\raisebox{-0.9ex}{\includegraphics[height=1.5em]{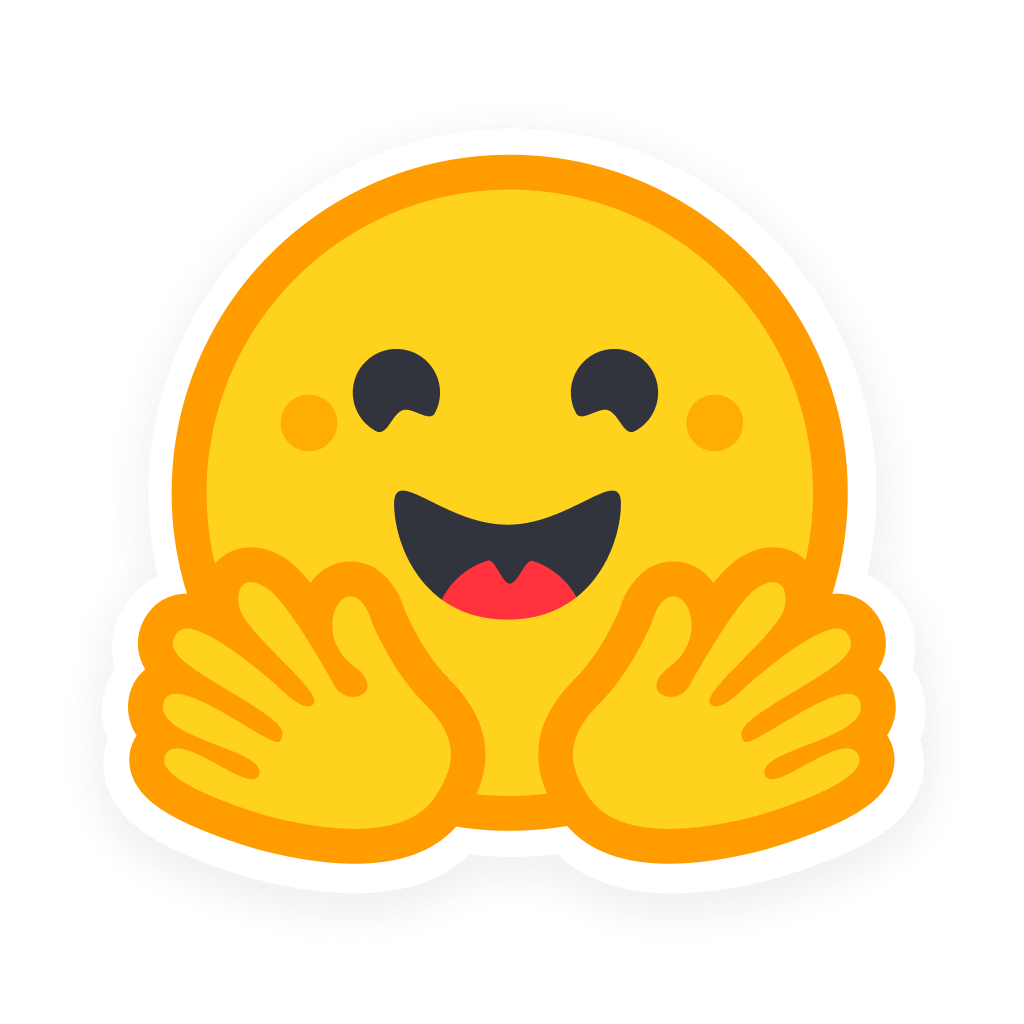}} #2}%
}

\newcommand{\gdrivelink}[2]{%
  \href{#1}{\raisebox{-0.3ex}{\includegraphics[height=1em]{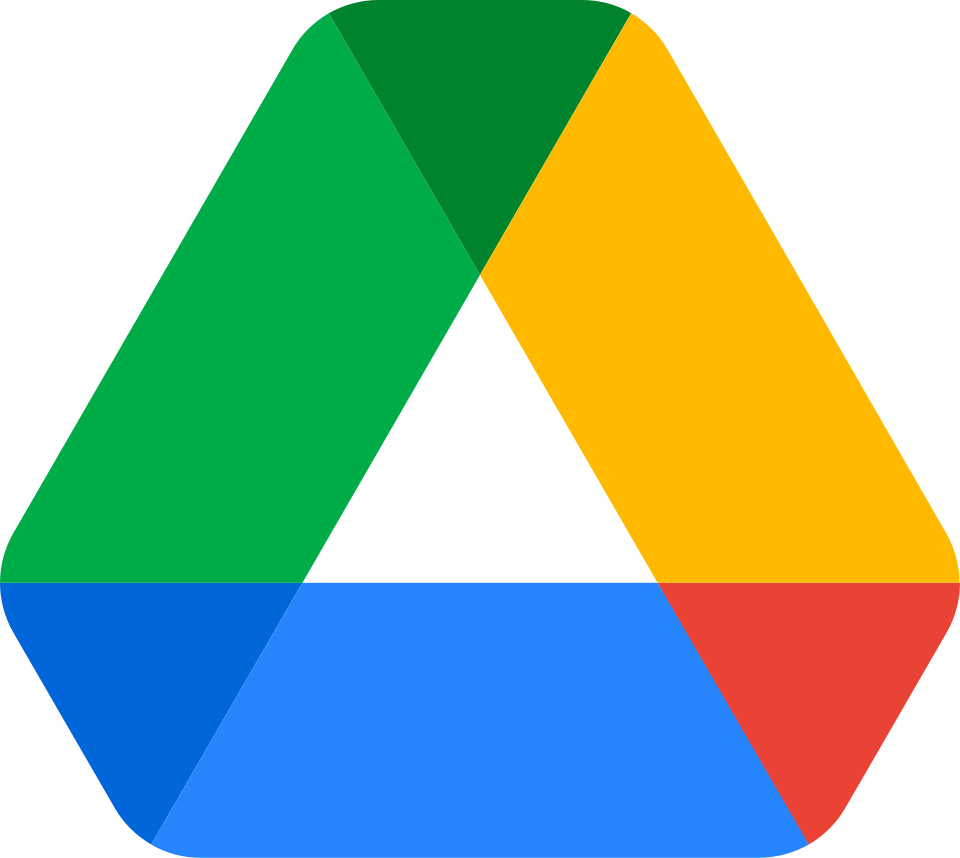}} #2}%
}
\newcommand{\gdrive}{\gdrivelink{https://drive.google.com/drive/folders/1iDfEV4RygpG4rCrb_TZ_FLu6Ur4qOdHH?usp=sharing}{KVzap predictions}}

\newcommand{\leaderboard}{\hflink{https://huggingface.co/spaces/nvidia/kvpress-leaderboard}{KVpress Leaderboard}\xspace}
\newcommand{\promptgreen}[1]{\textcolor{green!60!black}{\texttt{#1}}}
\newcommand{\promptred}[1]{\textcolor{red!70!black}{\texttt{#1}}}
\newcommand{\qwen}{Qwen3-8B\xspace}
\newcommand{\qwenb}{Qwen3-32B\xspace}
\newcommand{\llama}{Llama-3.1-8B-Instruct\xspace}

\title{KVzap: Fast, Adaptive, and Faithful KV Cache Pruning}

\author{
  Simon Jégou\protect\textsuperscript{*} Maximilian Jeblick \\
  \begin{center}
    \hflink{https://huggingface.co/collections/nvidia/kvzap}{NVIDIA/KVzap}\quad
    \githubrepo
  \end{center}
}

\begin{abstract}
\textbf{Abstract.} Growing context lengths in transformer-based language models have made the key-value (KV) cache a critical inference bottleneck. While many KV cache pruning methods have been proposed, they have not yet been adopted in major inference engines due to speed--accuracy trade-offs. We introduce KVzap, a fast, input-adaptive approximation of KVzip that works in both prefilling and decoding. On \qwen, \llama, and \qwenb across long-context and reasoning tasks, KVzap achieves $2$--$4\times$ KV cache compression with negligible accuracy loss and achieves state-of-the-art performance on the \leaderboard. Code and models are available at \githubrepo.
\end{abstract}

\begin{document}

\maketitle

{
  \renewcommand{\thefootnote}{\fnsymbol{footnote}} 
  \footnotetext[1]{Main contributor. Contact: \texttt{sjegou@nvidia.com}}
}

\begin{figure}[H]
    \centering
    \includegraphics[width=\linewidth]{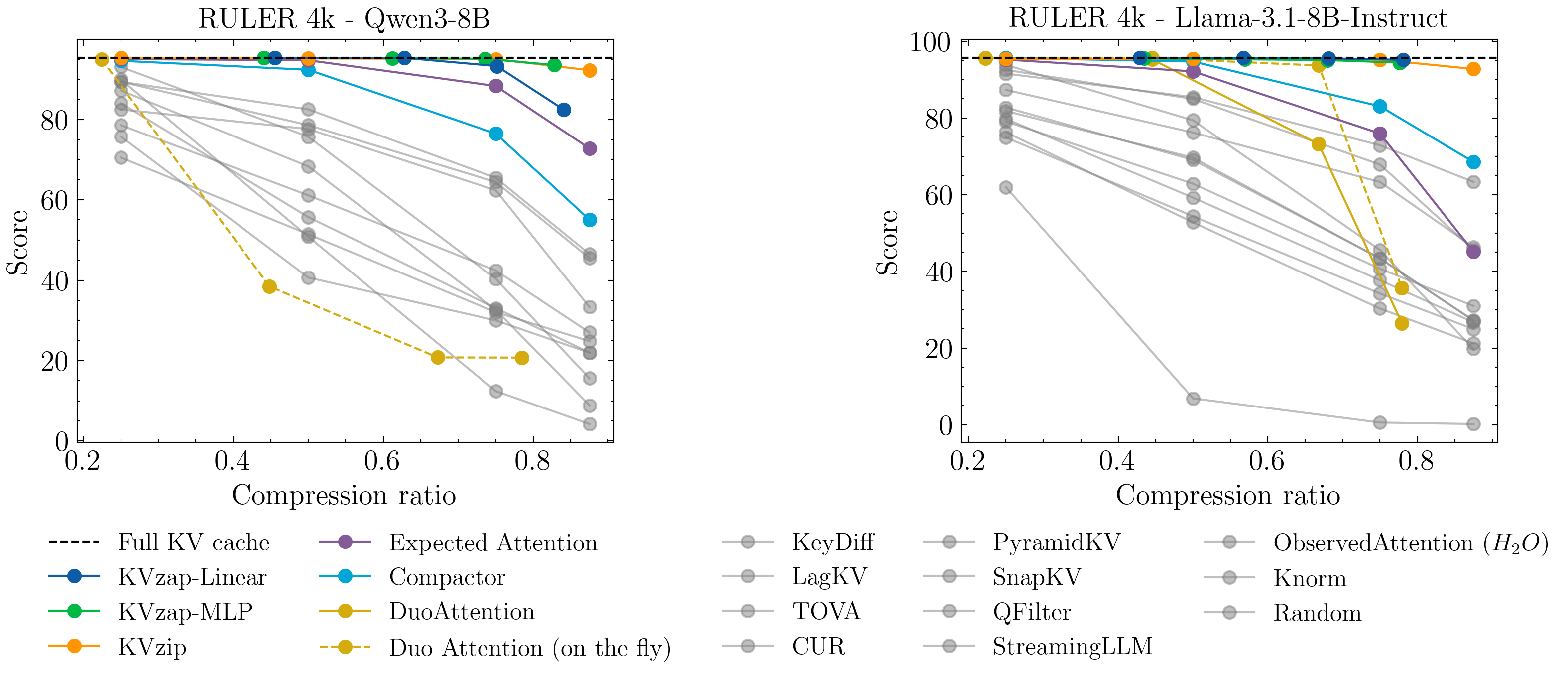}
    \caption{\leaderboard for \qwen (left) and \llama (right) comparing different KV cache pruning methods. The plots compare the accuracy on the RULER 4k dataset \citep{ruler} (y-axis) against the KV cache compression ratio (x-axis). KVzap achieves state-of-the-art performance on both models, matching KVzip \citep{kvzip} — which it approximates — while outperforming 15 other methods, including Expected Attention \citep{expected_attention}, Duo Attention \citep{duoattention}, and Compactor \citep{compactor}.}
    \label{fig:leaderboard}
\end{figure}

\section{Introduction}\label{sec:intro}

In transformer attention \citep{transformer}, each input token produces a set of key-value (KV) vector pairs that are stored in a cache and reused during autoregressive generation. The KV cache has shape $(2, L, H, T, D)$, where $L$ is the number of layers, $H$ the number of heads, $T$ the sequence length, and $D$ the key/value dimension. For example, in bfloat16 precision, the KV cache for a vanilla transformer like Llama1-65B \citep{llama1} ($L=80$, $H=64$, $D=128$) requires 335~GB of memory at $T=128$k. As sequence lengths grow to tens or hundreds of thousands of tokens, the KV cache becomes a dominant bottleneck for efficient LLM inference \citep{yaofu-long-context-challenge}, increasing GPU peak memory usage and time to first token while reducing decoding throughput.

Over the years, several architectural modifications have targeted specific axes of the KV cache to reduce its size. Along the $H$-axis, Grouped Query Attention (GQA \citep{gqa}) shares keys and values across multiple queries, yielding KV cache compression factors of 4$\times$ (Llama3 \citep{llama3}), 12$\times$ (GLM 4.5 \citep{glm45}), and up to 16$\times$ (Qwen3-235B-A22B \citep{qwen3}). Along the $D$-axis, DeepSeek V2 \citep{deepseekv2} introduces Multi-head Latent Attention (MLA) to perform a low-rank decomposition of keys and values, equivalent to a $4H/9$ compression. Along the $L$-axis, recent hybrid models interleave attention layers with sliding window attention (2$\times$ compression for GPT-OSS-120B \citep{gptoss}, 6$\times$ for Gemma3 \citep{gemma3}) or state space models (8$\times$ compression for Jamba \citep{jamba}, 4$\times$ compression for Kimi-Linear \citep{kimilinear}, 4.8$\times$ for Nemotron3 Nano \citep{nemotron3}).

Notably, no widely adopted architectural change compresses the KV cache along the $T$-axis. Sparse attention mechanisms, such as DSA in DeepSeek V3.2 \citep{DeepSeekV32}, retrieve only the most relevant KV pairs at each decoding step and can improve throughput, but they do not reduce the KV cache memory size.

Most attempts at KV cache compression on the $T$-axis rely on \textit{ad-hoc} pruning methods. Sparse attention is motivated by the idea that, within a head, attending to all past KV pairs is unnecessary for the \textit{next} decoding step; KV cache pruning goes further and assumes that some KV pairs will never be attended to \textit{at all}. \textcolor{gray}{Just as} a reader \textcolor{gray}{does} not pay equal attention to every word \textcolor{gray}{when understanding a sentence}, not all tokens are equally important, and some need not occupy $HL$ slots in KV cache memory. By retaining only KV pairs that are likely to be accessed, pruning methods can substantially reduce memory while preserving the information needed for faithful generation.

Pioneered by $H_2O$ \citep{h2o}, the \href{https://github.com/October2001/Awesome-KV-Cache-Compression}{{\faGithub}Awesome-KV-Cache-Compression} repository now lists dozens of KV cache pruning methods, with 20+ implemented in \githubrepo. Yet, none have been integrated into major inference engines such as vLLM \citep{vllm}, SGLang \citep{sglang}, or TRT-LLM \citep{trtllm}. We argue that a practical pruning method must meet the following criteria:

\begin{itemize}
    \item \textbf{Criterion 1: Fast and lightweight.} The pruning overhead must be negligible.
    \item \textbf{Criterion 2: Phase-agnostic.} The method must apply to both prefilling (long context) and decoding (reasoning tasks).
    \item \textbf{Criterion 3: Optimization-friendly.} The method must be compatible with kernels like FlashAttention2 \citep{flashattention2} or PagedAttention \citep{pagedattention}
    \item \textbf{Criterion 4: Faithful.} The method should cause minimal accuracy degradation on any task.
\end{itemize}

In this work, we introduce KVzap, a fast approximation of KVzip. Our contributions are the following:

\begin{itemize}
    \item We enhance the KVzip scoring with a normalization term inspired by \citep{expected_attention}, creating \textbf{KVzip+}.
    \item We demonstrate that KVzip+ scores can be approximated by a lightweight surrogate model trained on top of the model's hidden states.
    \item We introduce \textbf{KVzap}, a new KV cache pruning technique which applies these surrogate models to the hidden states, adopting DMS-style \citep{dms} inference: per-head compression, score-based pruning via thresholding, and delayed eviction with a sliding window.
\end{itemize}

\section{Related work}

\textbf{Block-Sparse Attention and Offloading.}
To mitigate the quadratic complexity of attention, recent research has explored architectural modifications and block-wise computation. Sparse attention mechanisms, such as DSA in DeepSeek V3.2 \citep{DeepSeekV32}, employ a lightweight indexer that computes coarse-grained similarity scores to retrieve only the most relevant KV blocks, reducing computational complexity to near-linear but requiring architectural support during pre-training. MoBA \citep{moba} partitions input sequences into independent segments, allowing only specific blocks to attend globally, which reduces prefilling latency but often requires fine-tuning to recover performance. SeerAttention \citep{seerattention} integrates hierarchical compression and selection directly into training. While these methods improve throughput, they focus on skipping computation rather than permanently reducing the memory footprint of the resident cache.

\textbf{Training-Free Eviction Methods.}
KV cache pruning methods try to identify KV pairs that are unlikely to be used by the attention mechanism and remove them, hence reducing memory while preserving the information needed for faithful generation. These methods typically apply eviction after attention, leaving the forward pass unchanged. Early approaches rely on heuristics to identify and evict less critical tokens without model retraining. $H_2O$ \citep{h2o} uses accumulated attention scores to retain ``heavy hitters'', and StreamingLLM \citep{streamingllm} keep recent tokens via sliding windows. While pioneering, these heuristics degrade significantly even at modest compression ratios. A variety of methods have since emerged \citep{snapkv, pyramidkv, scissorhands, fastgen}, though most fail to meet all of our criteria (see \href{https://github.com/October2001/Awesome-KV-Cache-Compression}{{\faGithub}Awesome-KV-Cache-Compression} for a comprehensive list). Many of these methods allow variable per-head compression rates, including AdaKV \citep{adakv} and Duo Attention \citep{duoattention}. Among these, KVzip \citep{kvzip} achieves the best performance in our benchmark (Figure~\ref{fig:leaderboard}) by defining token importance via ``context reconstruction'', retaining only the KV pairs necessary to reproduce the input. However, KVzip requires a computationally expensive double forward pass on an extended prompt, making it unsuitable for online inference (Criterion~1) and it cannot be applied during decoding (Criterion~2).

\textbf{End-to-End Learned Eviction.}
The most comparable approach to ours is Dynamic Memory Sparsification (DMS) \citep{dms}, which also learns a policy to dynamically prune the KV cache. DMS learns per-token eviction decisions via trainable weights applied to hidden states, using Gumbel-sigmoid reparametrization for differentiable training. The model is trained end-to-end to preserve its output distribution while masking KV pairs, with a loss combining KL divergence and a sparsity penalty. In contrast to KVzap, DMS evicts tokens before the attention pass, resulting in sparse prefilling. KVzap further differs in its training objective: it trains surrogate MLPs to approximate KVzip scores rather than using end-to-end distillation. We were unable to include DMS in our benchmark comparison due to the lack of publicly available checkpoints.

\textbf{Quantization.}
Orthogonal to token pruning, quantization methods such as KIVI \citep{kivi} and ZipCache \citep{zipcache} reduce KV cache memory by lowering numeric precision, demonstrating that KV pairs can be stored in INT4 or even INT2 formats with specialized kernels; these approaches can be combined with pruning for additional compression.

\section{Method}

\textbf{Summary:} In each transformer layer, KVzap applies a lightweight model to the input hidden states to predict importance scores and discards KV pairs whose score falls below a threshold $\tau$. The KVzap model is trained to approximate the scoring policy of an improved KVzip variant \citep{kvzip}.

\subsection{KVzip}

KVzip \citep{kvzip} currently stands as the state-of-the-art KV cache pruning method on the \leaderboard. While it reaches up to $4\times$ compression with minimal accuracy loss, it has major limitations that hinder adoption. First, it requires prefilling on an extended prompt twice as long as the input, making it prohibitively slow. Second, it cannot be used during decoding, which makes it unsuitable for reasoning tasks that generate thousands of tokens.

KVzip relies on a copy-and-paste pretext task to score the most important KV pairs. Given an input context \texttt{user:} \promptgreen{<prompt>} whose KV cache is to be compressed, it starts by building an extended prompt:

\begin{quote}
\texttt{user: }\promptgreen{<prompt>}\\
\texttt{Repeat the previous context exactly.}\\
\texttt{assistant: }\promptred{<prompt>}
\end{quote}

Then, for each head, the KV pair at position $i$ in the original \promptgreen{<prompt>} is scored as the maximum attention weight over the repeated \promptred{<prompt>} (and over heads in a group when GQA is used):

\begin{equation}
\label{eq:kvzip}
s_i = \max_{j \in \promptred{<prompt>}} a_{ji}
\end{equation}

Finally, the lowest-scoring KV pairs across heads and layers are removed. The intuition is that if, in a given head, the model pays little attention to position $i$ in the original \promptgreen{<prompt>} when repeating \promptred{<prompt>}, then the KV pair at $i$ carries little information and can be discarded.

\subsection{KVzip+} 

We enhance KVzip scoring by incorporating the analysis from \citep{expected_attention}. For a given transformer head at decoding step $j$, the hidden-state update is:

\begin{equation}
\label{eq:attention}
\mathbf{h}_j^{out} = \mathbf{h}_j + \sum_{i \le j} a_{ji} W_O \mathbf{v}_i
\end{equation}

where $\mathbf{h}_j$ is the input hidden state, $\mathbf{h}_j^{out}$ the output hidden state, $W_O$ the output projection matrix, and $\mathbf{v}_i$ the value vector. The term $a_{ji} W_O \mathbf{v}_i$ represents token $i$'s contribution to the residual stream $\mathbf{h}_j$. We incorporate this normalization into Eq.~\eqref{eq:kvzip} to define the KVzip+ score:

\begin{equation}
\label{eq:kvzip+}
s_i^+ = \max_{j \in \promptred{<prompt>}} a_{ji}\frac{\lVert W_O \mathbf{v}_i \rVert}{\lVert \mathbf{h}_j \rVert}
\end{equation}

\subsection{KVzap}

To address KVzip's limitations, we train a per-layer surrogate---either a linear layer or a two-layer MLP---to predict $H$ scores $\log(s^+)$ directly from the input hidden states $\mathbf{h}$, as proposed in DMS \citep{dms} (we use log-space to match the exponential nature of softmax attention). The model acts independently at each sequence position $t$: it maps $\mathbf{h}_t \in \mathbb{R}^{D_h}$ to scores in $\mathbb{R}^{H}$, where $D_h$ is the hidden dimension and $H$ is the number of KV heads. Because it uses only one or two matrix multiplications and depends only on hidden states, KVzap is computationally efficient (see Appendix \ref{sec:compute_overhead}) and can be applied during decoding. 

For training, we curate 1.2M pairs $(\mathbf{h}, \log(s^+))$ per KV head, sampled from \hflink{https://huggingface.co/datasets/nvidia/Nemotron-Pretraining-Dataset-sample}{Nemotron-Pretraining-Dataset-sample}. The dataset is diverse, covering English, multilingual, code, and mathematical text.

Another key difference lies in the eviction policy. Whereas KVzip enforces a fixed budget (e.g., keeping exactly 50\% of KV pairs), KVzap - like DMS - uses thresholding, discarding KV pairs whose predicted score falls below a fixed threshold $\tau$. Higher thresholds yield higher compression ratios. This makes KVzap input-adaptive: it dynamically adapts the compression rate based on the prompt information density, retaining more tokens for complex inputs and fewer for redundant ones.

Finally, to preserve local context, we keep a sliding window of the most recent $w=128$ tokens, following StreamingLLM \citep{streamingllm}. The full procedure is detailed in Algorithm \ref{code:kvzap}.

\begin{lstlisting}[style=github, numbers=none, caption={PyTorch pseudocode for KV cache pruning using KVzap during prefilling; similar to DMS \citep{dms} but applied after attention. Decoding is similar but needs a score buffer to enforce the sliding window}, label={code:kvzap}]
def compress(hidden_states, keys, values, kvzap_model, threshold, window=128):
    scores = kvzap_model(hidden_states)
    scores[..., -window:] = float("inf")
    indices = torch.where(scores >= threshold)
    return keys[indices], values[indices]
\end{lstlisting}

\section{Experiments}

All experiments were run on \qwen, \llama, and \qwenb and are fully reproducible via \githubrepo. Trained models are available in the \hflink{https://huggingface.co/collections/nvidia/kvzap}{NVIDIA/KVzap} collection, and full evaluation logs are provided in \gdrive.

\subsection{KVzap training} 

To generate training pairs $(\mathbf{h}, \log(s^+))$, we leveraged \hflink{https://huggingface.co/datasets/nvidia/Nemotron-Pretraining-Dataset-sample}{Nemotron-Pretraining-Dataset-sample}. The dataset contains 27k prompts split into 9 subsets (common crawl, multilingual, math, code, etc.). We filtered prompts to a length of 750--1,250 tokens to minimize the impact of sequence lengths on attention weights and then selected up to 500 prompts per subset for training and 5 for validation, resulting in roughly 2.4k prompts. We then randomly sampled 500 tokens per prompt to obtain 1.2M training pairs (per head), with 23k held out for validation.

For each KV head, we trained two types of surrogate models to predict $\log(s^+)$ from the hidden state $\mathbf{h}$: a linear model (\textbf{KVzap-Linear}) and a two-layer MLP (\textbf{KVzap-MLP}). The input dimension matches the model hidden size ($D_h$ = 4096 or 5120), and the output dimension is the number of KV heads ($H=8$). For MLPs, we used one hidden layer with width $D_h/8$ (512 or 640), followed by a GELU activation. In practice, KVzap-Linear and KVzap-MLP consist of a list of $L$ PyTorch modules \citep{pytorch}, with input size $(T, D_h)$ and output size $(T, H)$.

\begin{wraptable}{r}{0.45\textwidth}
    \centering
    \vspace{-12pt} 
    \caption{Average $R^2$ between KVzip+ scores and KVzap predictions on the validation set.}
    \label{tab:kvzap_r2}
    \begin{tabular}{lcc}
    \hline
    \textbf{Model} & \textbf{Linear} & \textbf{MLP} \\
    \hline
    \qwen  & 0.671 & \textbf{0.711} \\
    \llama & 0.743 & \textbf{0.772} \\
    \qwenb & 0.629 & \textbf{0.668} \\
    \hline
    \end{tabular}
    \vspace{-12pt}
\end{wraptable}

We report the average Squared Pearson correlation ($R^2$) over the $HL$ KV heads on the validation set in Table \ref{tab:kvzap_r2}. Both surrogates reach $R^2$ in the $0.60$--$0.80$ range, showing that the expensive KVzip+ score can be approximated from hidden states. Across all models, KVzap-MLP consistently outperforms KVzap-Linear. A more detailed analysis is provided in Appendix \ref{sec:kvzap_training}.

\subsection{Compute and memory overhead}

KVzap adds negligible overhead: across all models, its relative compute cost is bounded by $1.1\%$ for KVzap-MLP and $0.02\%$ for KVzap-Linear when considering linear projections \textit{only} (Table \ref{tab:kvzap_overhead} in Appendix \ref{sec:compute_overhead}). The relative memory overhead matches these bounds, and in long-context regimes the quadratic attention cost dominates, making KVzap's overhead negligible. Finally, during decoding---which is strictly memory-bandwidth bound---KVzap’s additional FLOPs effectively utilize idle GPU cycles that would otherwise be stalled by KV cache retrieval \citep{memorygap}.

\subsection{Prefilling and decoding tasks}

KV cache pruning is most impactful for tasks involving thousands of tokens, during prefilling (long inputs) or decoding (long outputs). To assess KVzap across these regimes, we evaluated KVzap-Linear and KVzap-MLP on two long-context benchmarks---RULER \citep{ruler} ($n=6500$) and LongBench \citep{longbench} ($n=4750$)---and one reasoning benchmark, AIME25 \citep{aime25} ($n=30$).

\paragraph{Experimental Setup} We evaluate KVzap using thresholds $\tau \in \{-6, -5, -4, -3\}$ for \qwen and \qwenb, and $\tau \in \{-9, -8, -7, -6\}$ for \llama. For RULER and LongBench, we used greedy decoding and disabled reasoning; for AIME25, we evaluated \qwen and \qwenb models with reasoning and sampling parameters recommended in the Qwen3 model card ($\text{temperature}=0.6$, top-$p=0.95$, top-$k=20$). In all experiments, KV cache compression was applied after the attention operation. 

\subsection{RULER}

RULER \citep{ruler} evaluates long-context capabilities across four task categories---retrieval, multi-hop tracing, aggregation, and question answering---over 13 subsets with sequence lengths ranging from 4k to 128k.

On RULER 4k, KVzap achieves state-of-the-art results for both \qwen and \llama (Figure \ref{fig:leaderboard}), significantly outperforming 15 concurrent KV cache pruning methods.

\begin{figure}[t]
    \centering
    \includegraphics[width=\linewidth]{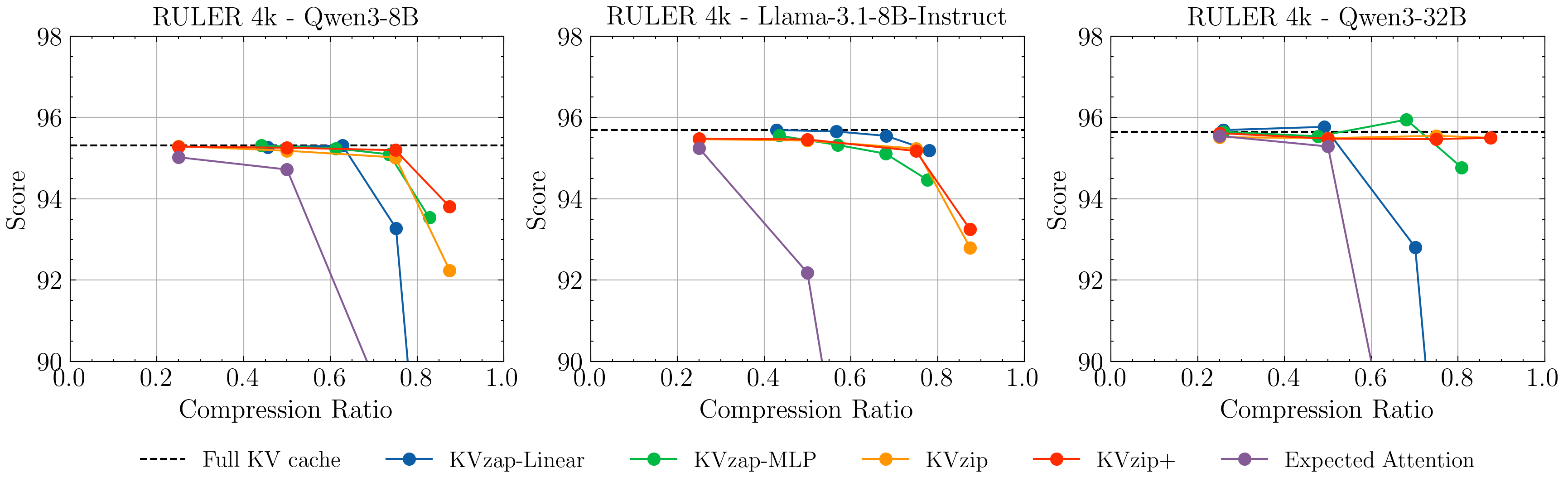}
    \caption{\textbf{RULER 4k results} for \qwen (left), \llama (middle), and \qwenb (right). Zoomed-in view (y-axis range [90, 100]) of the high-performance region from Figure \ref{fig:leaderboard}. KVzap surrogates perform comparably to---and sometimes exceed---the KVzip+ oracle they approximate.}
    \label{fig:ruler_4k}
\end{figure}

We provide a magnified view in Figure \ref{fig:ruler_4k}, comparing KVzap variants against KVzip, KVzip+, and Expected Attention \citep{expected_attention}. A few trends emerge: (1) KVzip+ consistently matches or exceeds KVzip, validating our normalization; (2) KVzap maintains perfect accuracy up to $3$--$4\times$ compression; (3) For Qwen models, KVzap-MLP outperforms KVzap-Linear, which degrades sharply at high compression; (4) Surprisingly, KVzap-Linear excels on \llama despite lower $R^2$ than KVzap-MLP and even outperforms the KVzip+ oracle it approximates.

\subsection{LongBench} 

\begin{figure}[t]
    \centering
    \includegraphics[width=\linewidth]{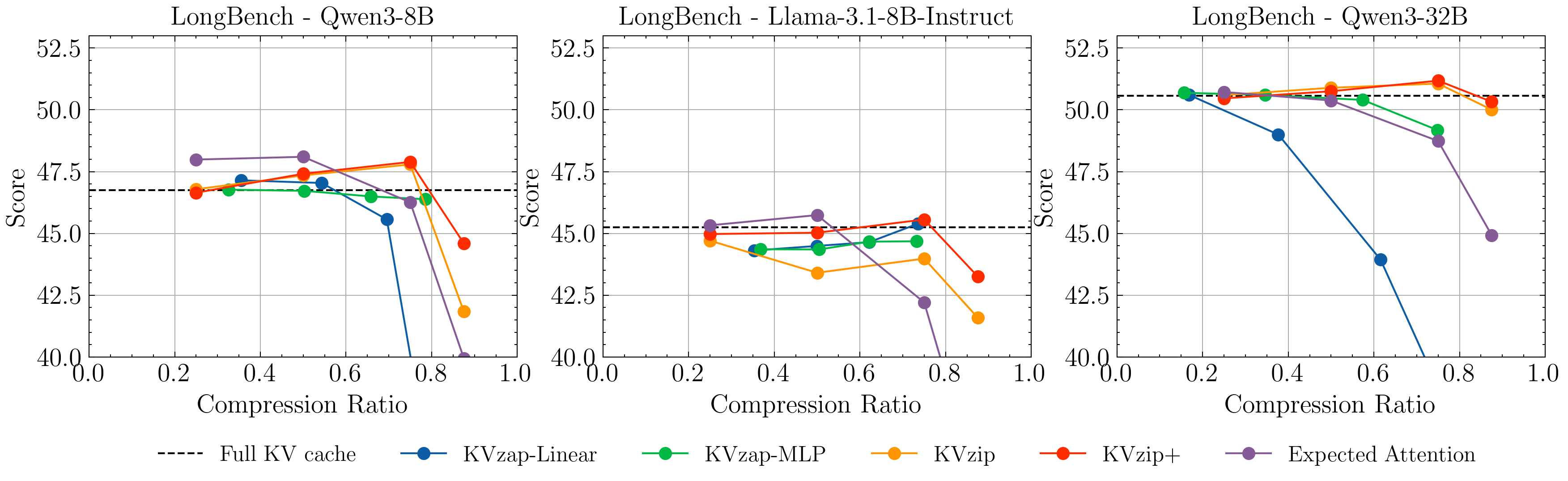}
    \caption{\textbf{LongBench results} for \qwen (left), \llama (middle), and \qwenb (right). KVzap models again maintain accuracy close to the full KV cache baseline. The elevated scores for Expected Attention are primarily driven by outliers in TREC, one of the 21 subsets of LongBench; see Figure \ref{fig:longbench_no_trec} for results excluding TREC.}
    \label{fig:longbench}
\end{figure}

LongBench \citep{longbench} evaluates long-context capabilities across six task categories---single-document QA, multi-document QA, summarization, few-shot learning, synthetic tasks, and code completion---spanning 21 subsets in English and Chinese. We report the average performance across subsets in Figure \ref{fig:longbench}.

Mirroring the main RULER results, (1) KVzip+ consistently matches or outperforms KVzip, and (2) KVzap models maintain near-perfect accuracy up to 2–3$\times$ compression. Notably, the same thresholds $\tau$ yield lower compression ratios, likely due to data characteristics: RULER samples are synthetic and repetitive, whereas LongBench consists mostly of real-world data with higher information density.

At first glance, Expected Attention \citep{expected_attention} appears to surpass the full KV cache baseline for \qwen and \llama at lower compression ratios. A closer look reveals this is largely driven by outlier accuracies on the TREC subset (see Figure \ref{fig:longbench_no_trec}, \ref{fig:longbench_details_qwen} and \ref{fig:longbench_details_llama}), and that Expected Attention degrades on several subsets where KVzap stays close to the full KV cache baseline. Higher TREC accuracy at high compression may be explained by the over-prompting phenomenon \citep{overprompting}: in a few-shot learning task like TREC, adding more examples can counter-intuitively reduce accuracy.

The generally low accuracy on most LongBench subsets, combined with their small size (typically $n=200$), leads to high variance, making results harder to interpret conclusively.

\subsection{AIME25} 

\begin{figure}[t]
    \centering
    \includegraphics[width=0.8\linewidth]{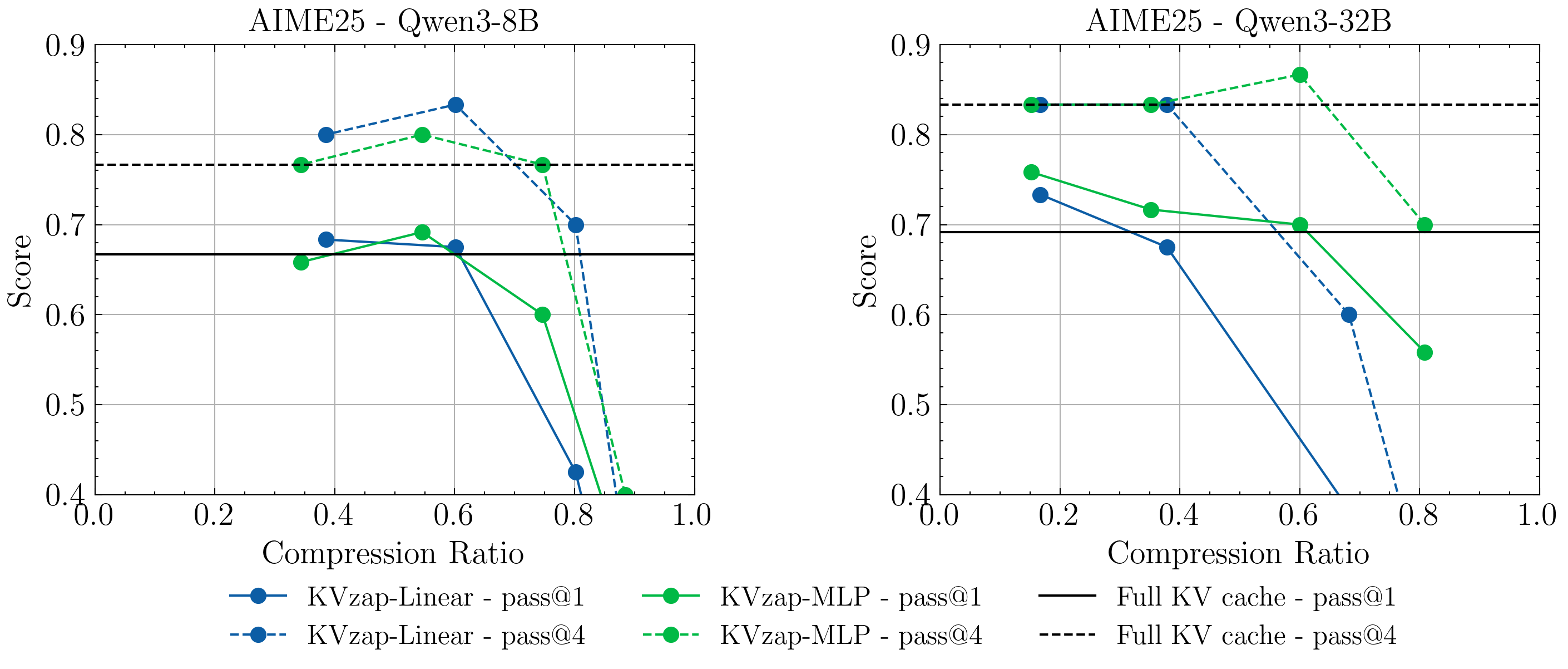}
    \caption{\textbf{AIME25 Reasoning Performance.} Comparison of pass@1 (solid lines) and pass@4 (dashed lines) accuracy for \qwen (left) and \qwenb (right). KVzap-MLP maintains robust performance even when discarding over 50\% of the KV cache.}
    \label{fig:aime25}
\end{figure}

The AIME25 benchmark \citep{aime25} consists of 30 Olympiad-level, integer-answer problems from the 2025 American Invitational Mathematics Examination. We evaluated KVzap with 4 rollouts per question, a generation limit of 32k tokens, and we report average pass@1 and pass@4 in Figure \ref{fig:aime25}. KVzap-MLP preserves reasoning accuracy even at compression ratios exceeding $2\times$.

\subsection{Adaptive compression}

Figures~\ref{fig:ruler_4k}, \ref{fig:longbench}, and \ref{fig:aime25} show that the maximum compression that does not degrade accuracy is task-dependent (e.g., higher on RULER and lower on LongBench). KVzap's thresholding captures this automatically: the same threshold $\tau$ translates into different compression ratios across benchmarks.

Table~\ref{tab:summary_results} reports the best KVzap configuration (Linear/MLP and $\tau$) per model. Overall, KVzap achieves $2.7$--$3.5\times$ average KV cache compression while maintaining accuracy across model scales and tasks.

\begin{table}[t!]
\centering
\caption{Performance of KVzap across models and datasets. Arrows ($\rightarrow$) show the change from full to compressed KV cache. Values in parentheses indicate the KV cache compression ratio (removed fraction). For each model, we report the best KVzap configuration (Linear/MLP and threshold $\tau$).}
\label{tab:summary_results}
\setlength{\tabcolsep}{6pt}
\begin{tabular}{lccc}
\toprule
 & \textbf{Qwen3-8B} & \textbf{Llama-3.1-8B} & \textbf{Qwen3-32B} \\
\midrule
KVzap model & MLP & Linear & MLP \\
Parameters  & 76M & 1.1M   & 210M \\
Threshold   & $\tau=-4$ & $\tau=-7$ & $\tau=-4$ \\
\midrule
RULER 4k &  $95.32 \rightarrow 95.09$ \small{($0.74$)} &  $95.69 \rightarrow 95.55$ \small{($0.68$)} &  $95.65 \rightarrow 95.95$ \small{($0.68$)} \\
\addlinespace
RULER 16k &  $92.99 \rightarrow 92.78$ \small{($0.72$)} &  $93.42 \rightarrow 93.29$ \small{($0.70$)} &  $95.19 \rightarrow 94.96$ \small{($0.65$)} \\
\addlinespace
LongBench &  $46.74 \rightarrow 46.49$ \small{($0.66$)} &  $45.25 \rightarrow 44.65$ \small{($0.62$)} &  $50.56 \rightarrow 50.40$ \small{($0.57$)} \\
\addlinespace
AIME25 (pass@4) &  $0.77 \rightarrow 0.77$ \small{($0.75$)} & -- &  $0.83 \rightarrow 0.87$ \small{($0.60$)} \\
\midrule
Average compression ratio &  $0.72$ ($3.5\times$) &  $0.67$ ($3.0\times$) &  $0.63$ ($2.7\times$) \\
\bottomrule
\end{tabular}
\end{table}

\subsection{Ablations}

\paragraph{Threshold-based pruning}

\begin{figure}[t]
    \centering
    \includegraphics[width=\linewidth]{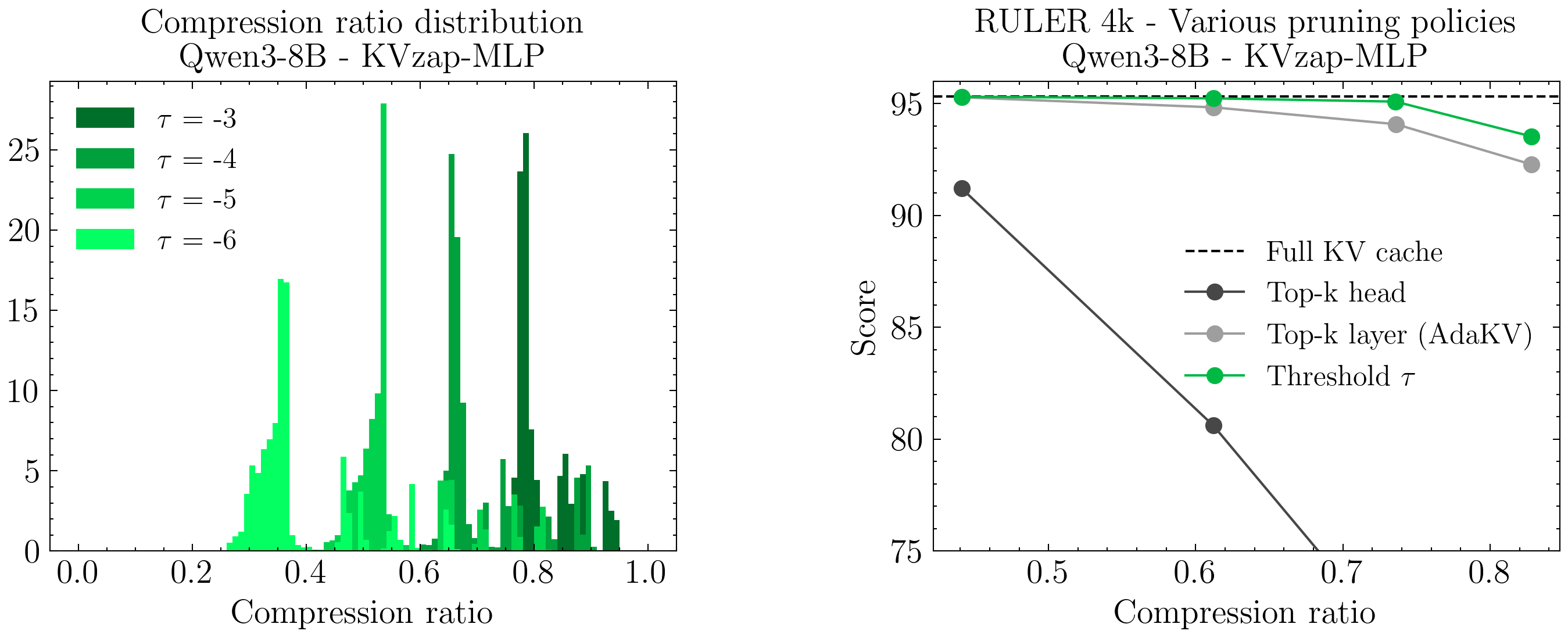}
    \caption{Distribution of compression ratios for \qwen and KVzap-MLP on RULER 4k, LongBench, and AIME25 (left), and comparison to an alternative pruning method (right).}
    \label{fig:cr_distribution}
\end{figure}

KVzap uses score thresholding rather than fixed top-$k$ selection, allowing the compression rate to adapt to prompt complexity. Figure \ref{fig:cr_distribution} (left) highlights this adaptability, showing up to 20\% variation across prompts. As shown in Figure \ref{fig:cr_distribution} (right), thresholding outperforms fixed-ratio top-$k$ selection, whether per-head or per-layer (AdaKV \citep{adakv}).

\paragraph{Sliding Window}
We analyze the impact of sliding-window size $w$ on LongBench-LCC with \qwen, KVzap-MLP, and $\tau=-4$. Without a local window ($w=0$), accuracy drops to 28.37\% because the input hidden states do not explicitly encode position information. Enforcing $w=128$ restores performance to 62.51\%, while increasing to $w=512$ yields no additional gain (62.37\%).

\section{Discussion}

Across multiple models (\qwen, \llama, \qwenb) and benchmarks (RULER, LongBench, AIME25), we show that KVzap achieves $2$--$4\times$ KV cache compression with negligible accuracy loss. Its design---a lightweight linear or MLP model applied to hidden states---is computationally efficient and easy to integrate. Still, limitations and future directions remain.

\paragraph{Scope and Generalization}
First, while results on a 32B model are encouraging, further validation is needed on larger open-source models (e.g., GLM 4.7 \citep{glm45}, Qwen3-235B-A22B \citep{qwen3}) and architectures with sparse attention (e.g., DeepSeek V3.2 \citep{DeepSeekV32}). Evaluation could also be extended to more reasoning benchmarks, agentic tasks, and short-context knowledge tasks.

\paragraph{Ad-hoc vs. End-to-End Training}
Second, KVzap is not training-free, and like most KV cache pruning methods, it is a post-hoc addition. In the long run, end-to-end integration often prevails in deep learning, much as Multi-Token Prediction \citep{mtp} is superseding ad-hoc speculative decoding techniques such as Medusa \citep{medusa}. Although still rare, end-to-end pruning objectives like DMS \citep{dms} exist and may eventually yield better performance. Nonetheless, KVzap provides further evidence that LLMs do not fully exploit the KV cache and that unused KV pairs can be easily identified from hidden states.

\paragraph{Implementation Challenges}
Third, turning compression into wall-clock speedups and GPU memory savings requires careful engineering and was not explored here. KVzap introduces non-uniform cache lengths across heads, requiring PagedAttention kernels \citep{pagedattention} that handle variable-length blocks. Prior work such as DMS \citep{dms}, Compactor \citep{compactor}, AdaKV \citep{adakv}, have shown this is feasible, but kernel optimization remains non-trivial. Since KVzap relies only on hidden states, pruning could also be applied before attention to directly accelerate prefilling.

\paragraph{Conclusion}

Despite these challenges, we believe KVzap's combination of simplicity, high compression ratios, and robust performance across tasks and models makes it a prime candidate for production deployment, potentially bridging the gap between academic pruning research and real-world inference engines.

\section*{Acknowledgments}
We thank Alessio Devoto for his careful reading of the manuscript and for providing detailed and constructive feedback.

\bibliographystyle{style}
\bibliography{references}

\clearpage
\appendix

\section{KVzap model training}\label{sec:kvzap_training}

\begin{figure}[t!]
    \centering
    \includegraphics[width=\linewidth]{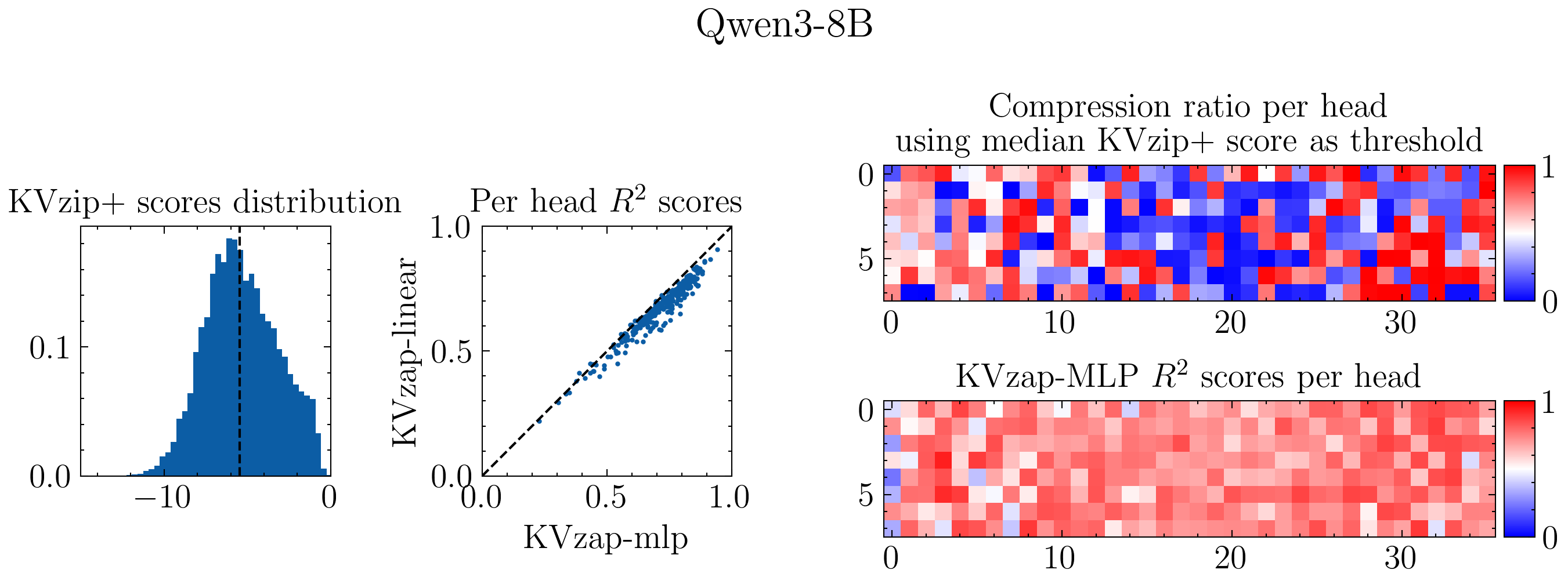}
    \caption{Detailed KVzap evaluation analysis for \qwen. \textbf{Left:} Distribution of KVzip+ scores computed on the 23k validation pairs. \textbf{Middle:} Correlation between the $R^2$ performance of KVzap-MLP (x-axis) and KVzap-Linear (y-axis) for each head. \textbf{Upper Right:} Fraction of KV pairs falling below the median KVzip+ score. \textbf{Lower Right:} Heatmap of $R^2$ scores for KVzap-MLP across all heads.}
    \label{fig:qwen_kvzap_scores}
\end{figure}

\begin{figure}[t!]
    \centering
    \includegraphics[width=\linewidth]{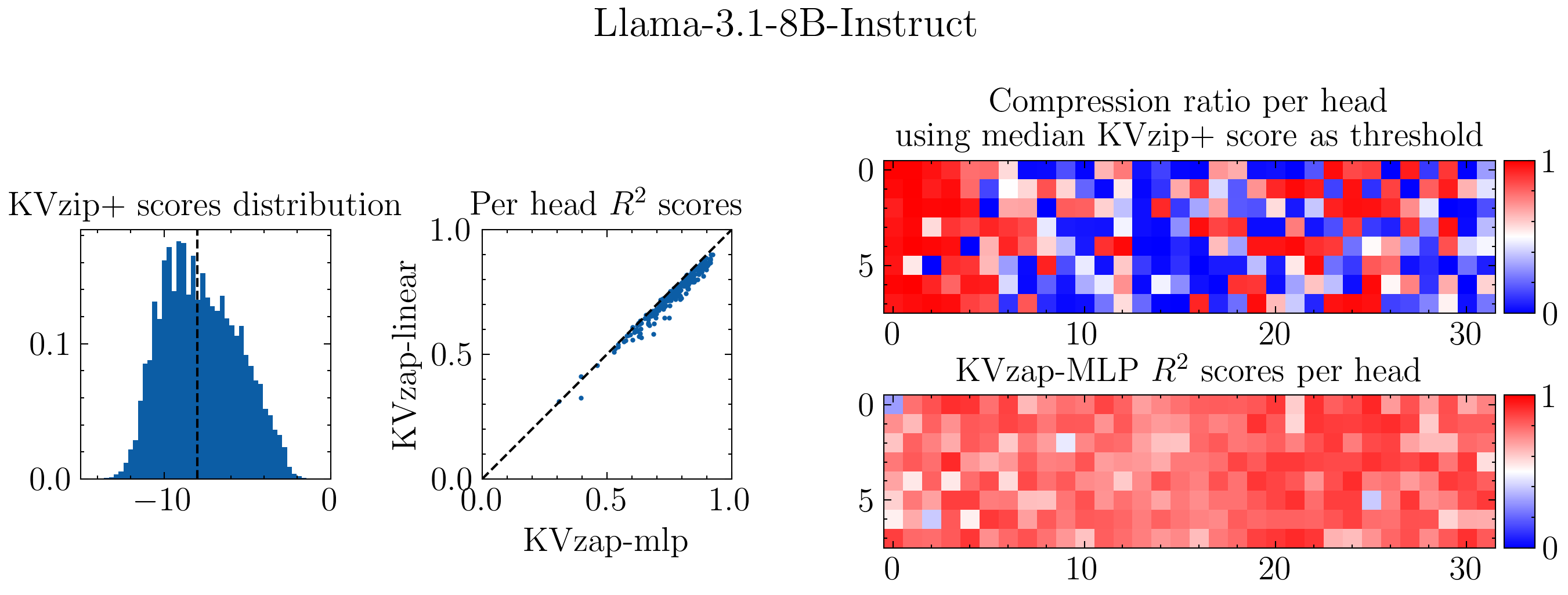}
    \caption{Detailed KVzap evaluation analysis for \llama. \textbf{Left:} Distribution of KVzip+ scores computed on the 23k validation pairs. \textbf{Middle:} Correlation between the $R^2$ performance of KVzap-MLP (x-axis) and KVzap-Linear (y-axis) for each head. \textbf{Upper Right:} Fraction of KV pairs falling below the median KVzip+ score. \textbf{Lower Right:} Heatmap of $R^2$ scores for KVzap-MLP across all heads.}
    \label{fig:llama_kvzap_scores}
\end{figure}

\begin{figure}[t!]
    \centering
    \includegraphics[width=\linewidth]{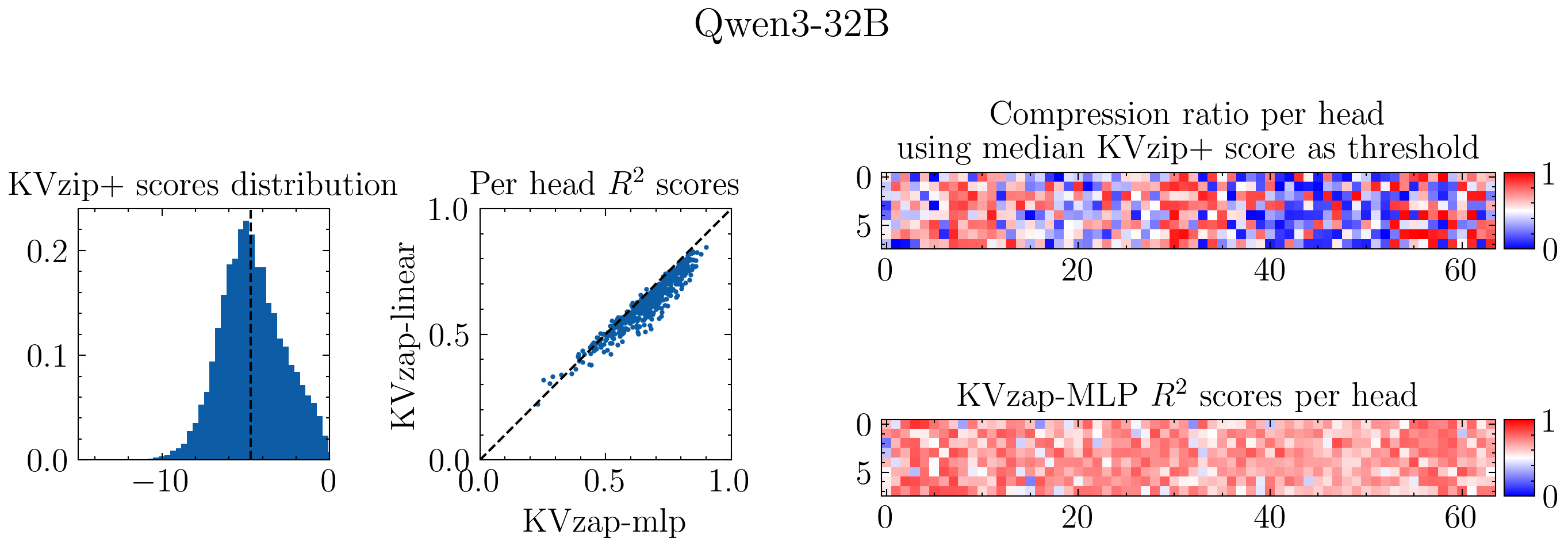}
    \caption{Detailed KVzap evaluation analysis for \qwenb. \textbf{Left:} Distribution of KVzip+ scores computed on the 23k validation pairs. \textbf{Middle:} Correlation between the $R^2$ performance of KVzap-MLP (x-axis) and KVzap-Linear (y-axis) for each head. \textbf{Upper Right:} Fraction of KV pairs falling below the median KVzip+ score. \textbf{Lower Right:} Heatmap of $R^2$ scores for KVzap-MLP across all heads.}
    \label{fig:qwenb_kvzap_scores}
\end{figure}

We report detailed distributions of KVzip+ and $R^2$ scores in Figures \ref{fig:qwen_kvzap_scores} (\qwen), \ref{fig:llama_kvzap_scores} (\llama), and \ref{fig:qwenb_kvzap_scores} (\qwenb).

The \llama score distribution is significantly lower than for \qwen and \qwenb, motivating lower pruning thresholds in our experiments. Across all models, KVzap-MLP consistently achieves higher $R^2$ than KVzap-Linear. Both surrogates perform worse in the first transformer layer, suggesting that KVzip+ scores are harder to infer from token embeddings alone. Predicting scores directly from keys and values $(\mathbf{k}, \mathbf{v})$ instead of hidden states $\mathbf{h}$ resulted in strictly lower $R^2$. We also acknowledge a potential train-test distribution shift as KVzap is trained on prompts limited to 1,250 tokens.

We trained KVzap-Linear using scikit-learn \citep{scikit-learn} and KVzap-MLP using skorch \citep{skorch}. Future work could further improve accuracy through better data selection and hyperparameter tuning.

\section{KVzap compute and memory overhead}\label{sec:compute_overhead}

We analyze the compute overhead introduced by KVzap within a single transformer decoder layer, relative to the cost of all linear projections in the layer: the attention projection matrices ($W_Q$, $W_K$, $W_V$, $W_O$) and the feed-forward network (FFN). We ignore the quadratic attention matrix multiplication and nonlinearities, yielding a conservative upper bound on the relative compute cost.

Assuming a GQA setting with $H_Q$ query heads, $H$ key-value heads, head dimension $D$, hidden dimension $D_h$, and a SwiGLU FFN intermediate dimension $D_{\text{int}}$, the FLOPs from linear projections are:

\begin{equation}
C = C_{\text{attn}} + C_{\text{ffn}}
= 4 D_h \big( H_Q D + H D \big) + 6 D_h D_{\text{int}},
\end{equation}

We compare against KVzap-MLP, consisting of two linear layers $W_1 \in \mathbb{R}^{D_h \times D_h/8}$ and $W_2 \in \mathbb{R}^{D_h/8 \times H}$:
\begin{equation}
C_{\text{KVzap-MLP}} = 2 \left( D_h \cdot \frac{D_h}{8} \right) + 2 \left( \frac{D_h}{8} \cdot H \right) = \frac{D_h}{4}(D_h + H)
\end{equation}
and KVzap-Linear, consisting of a single projection from $D_h$ to $H$:
\begin{equation}
C_{\text{KVzap-Linear}} = 2 D_h H
\end{equation}

\begin{table}[H]
\centering
\small
\begin{tabular}{lccccccc}
\toprule
Model
& $H_Q$ & $H$ & $D$ & $D_h$ & $D_{\text{int}}$
& $\frac{C_{\text{KVzap-MLP}}}{C}$
& $\frac{C_{\text{KVzap-Linear}}}{C}$ \\
\midrule
Qwen3-8B
& 32 & 8 & 128 & 4096 & 12288
& 1.09\% & 0.02\% \\
Llama-3.1-8B-Instruct
& 32 & 8 & 128 & 4096 & 14336
& 0.96\% & 0.02\% \\
Qwen3-32B
& 64 & 8 & 128 & 5120 & 25600
& 0.67\% & 0.01\% \\
\bottomrule
\end{tabular}
\caption{Relative compute overhead of KVzap compared to a single transformer layer, considering only linear projections.}
\label{tab:kvzap_overhead}
\end{table}

Table~\ref{tab:kvzap_overhead} reports the resulting relative compute overhead for \qwen, \llama, and \qwenb, showing a maximum overhead of $1.1\%$ for KVzap-MLP and $0.02\%$ for KVzap-Linear. In long-context regimes, the quadratic cost of attention dominates the overall complexity, making this overhead effectively negligible.

The relative memory overhead (ignoring biases) matches the compute overhead, as the factor of two introduced in FLOPs counting cancels out in the ratio.

Overall, KVzap’s additional parameters introduce no meaningful memory or compute overhead.

\section{Detailed benchmark results}\label{sec:detailed_results}

\paragraph{RULER}

\begin{figure}[t]
    \centering
    \includegraphics[width=\linewidth]{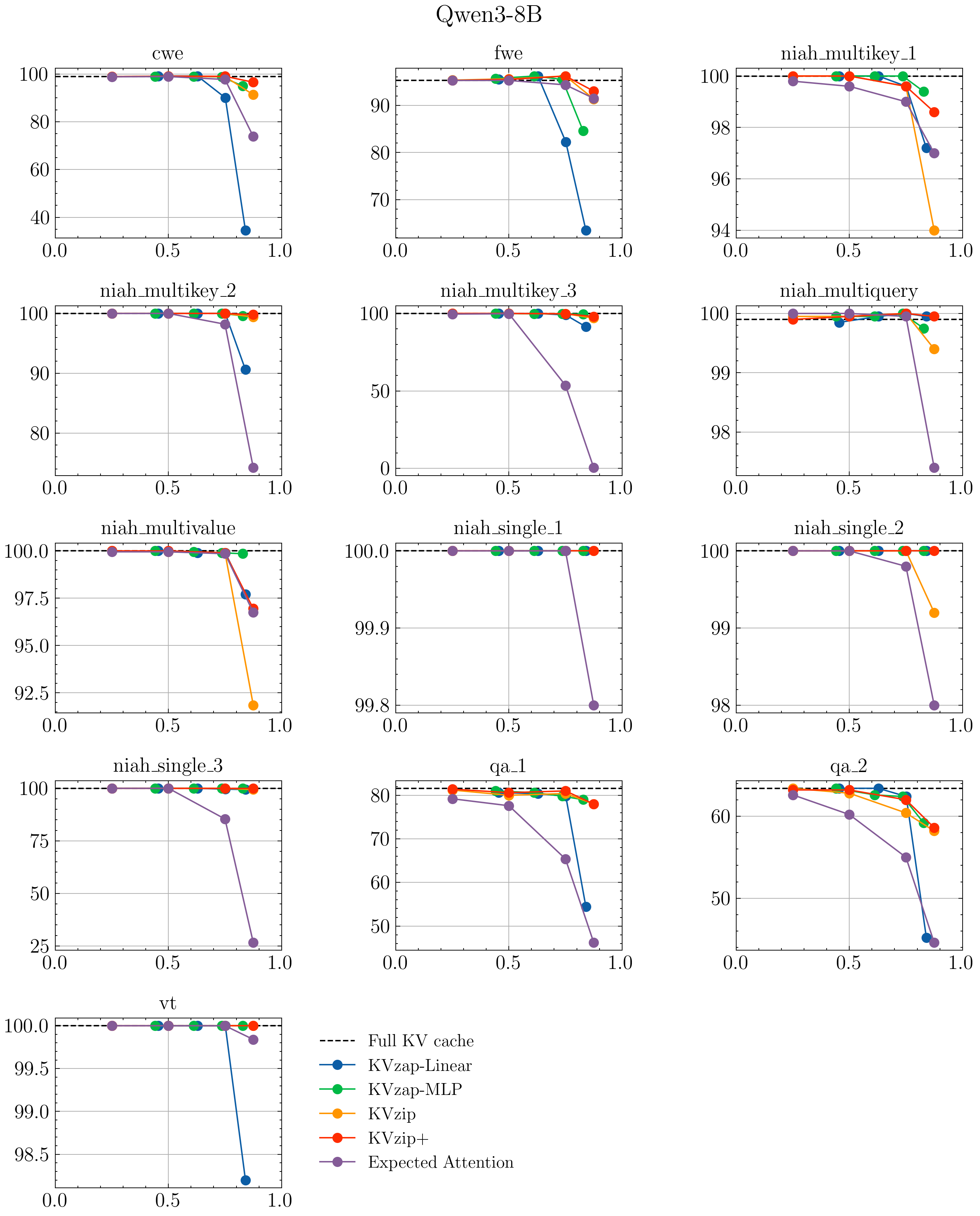}
    \caption{RULER 4k results for \qwen on each of the 13 subsets}
    \label{fig:ruler_4k_details_qwen}
\end{figure}

\begin{figure}[t]
    \centering
    \includegraphics[width=\linewidth]{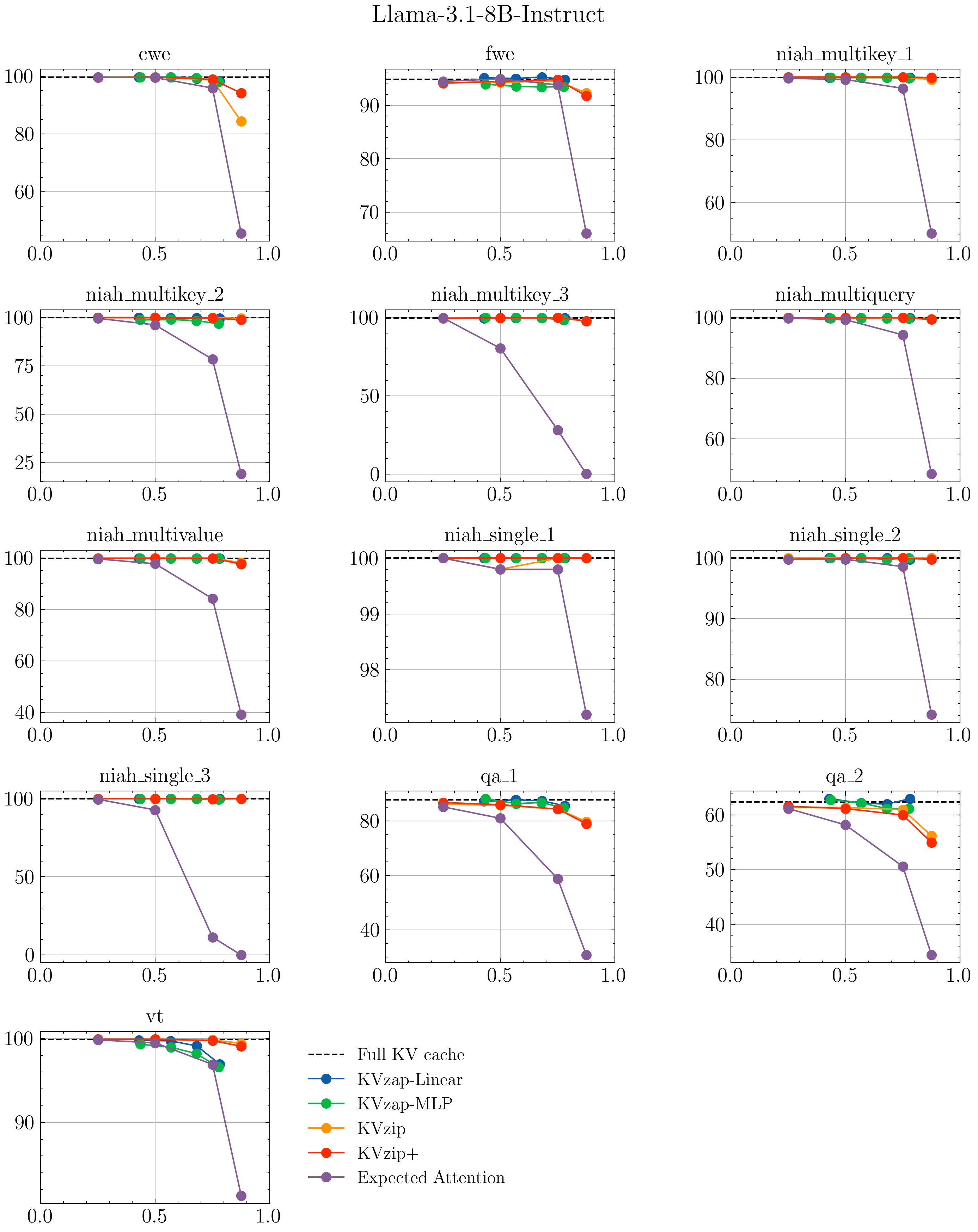}
    \caption{RULER 4k results for \llama on each of the 13 subsets}
    \label{fig:ruler_4k_details_llama}
\end{figure}

\begin{figure}[t]
    \centering
    \includegraphics[width=\linewidth]{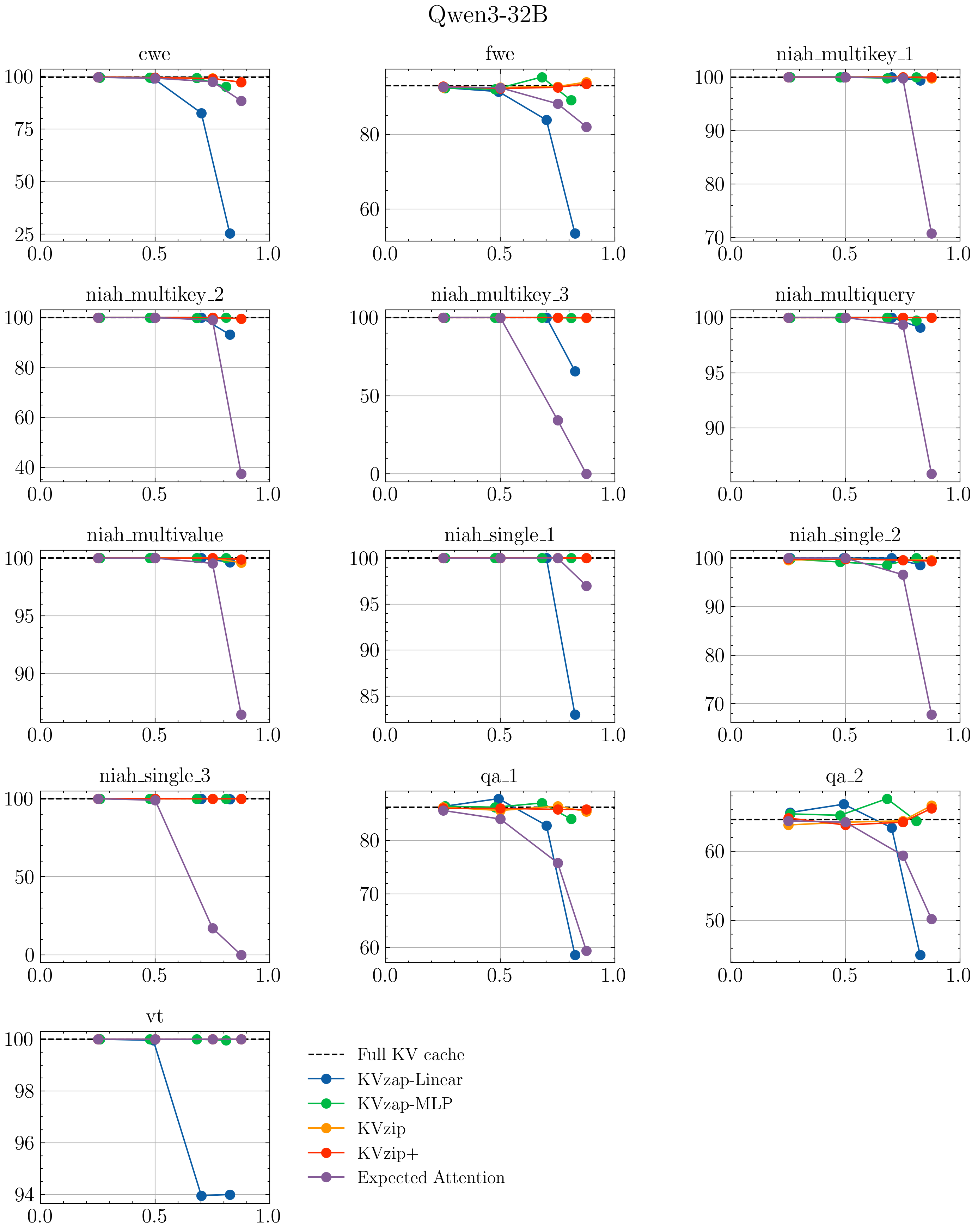}
    \caption{RULER 4k results for \qwenb on each of the 13 subsets}
    \label{fig:ruler_4k_details_qwenb}
\end{figure}

Figures \ref{fig:ruler_4k_details_qwen}--\ref{fig:ruler_4k_details_qwenb} provide per-subset results for RULER 4k (\qwen, \llama, \qwenb).

\paragraph{LongBench}

\begin{figure}[t]
    \centering
    \includegraphics[width=\linewidth]{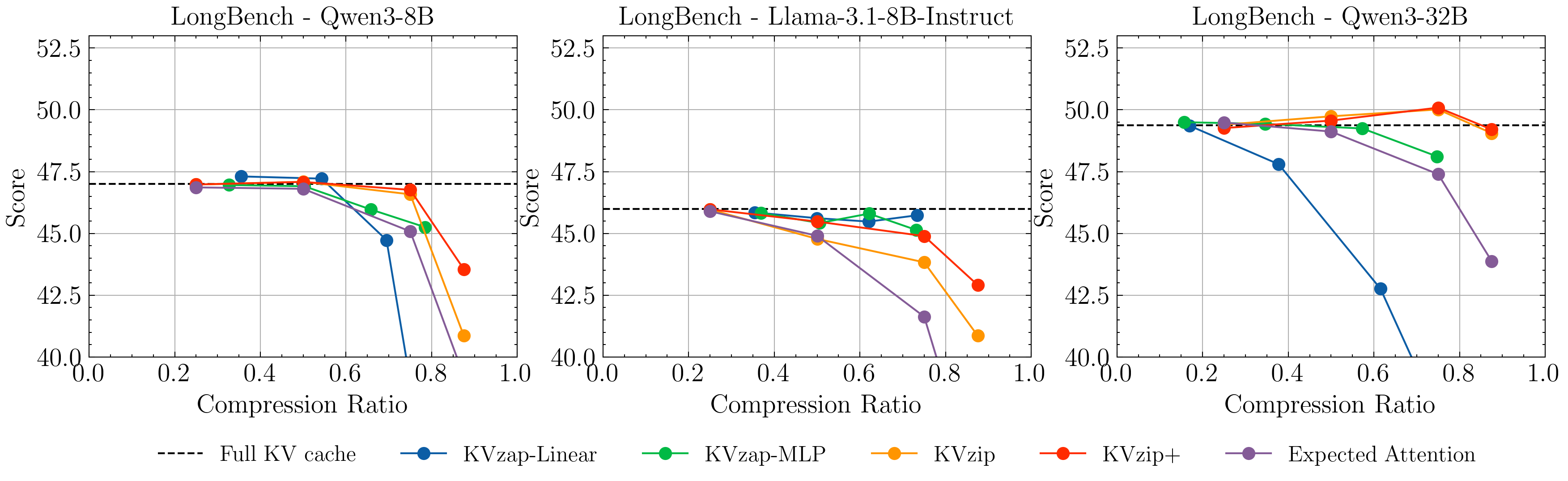}
    \caption{\textbf{LongBench results} for \qwen (left), \llama (middle), and \qwenb (right). Average score across 20/21 subsets, after excluding the TREC subset.}
    \label{fig:longbench_no_trec}
\end{figure}

\begin{figure}[t]
    \centering
    \includegraphics[width=\linewidth]{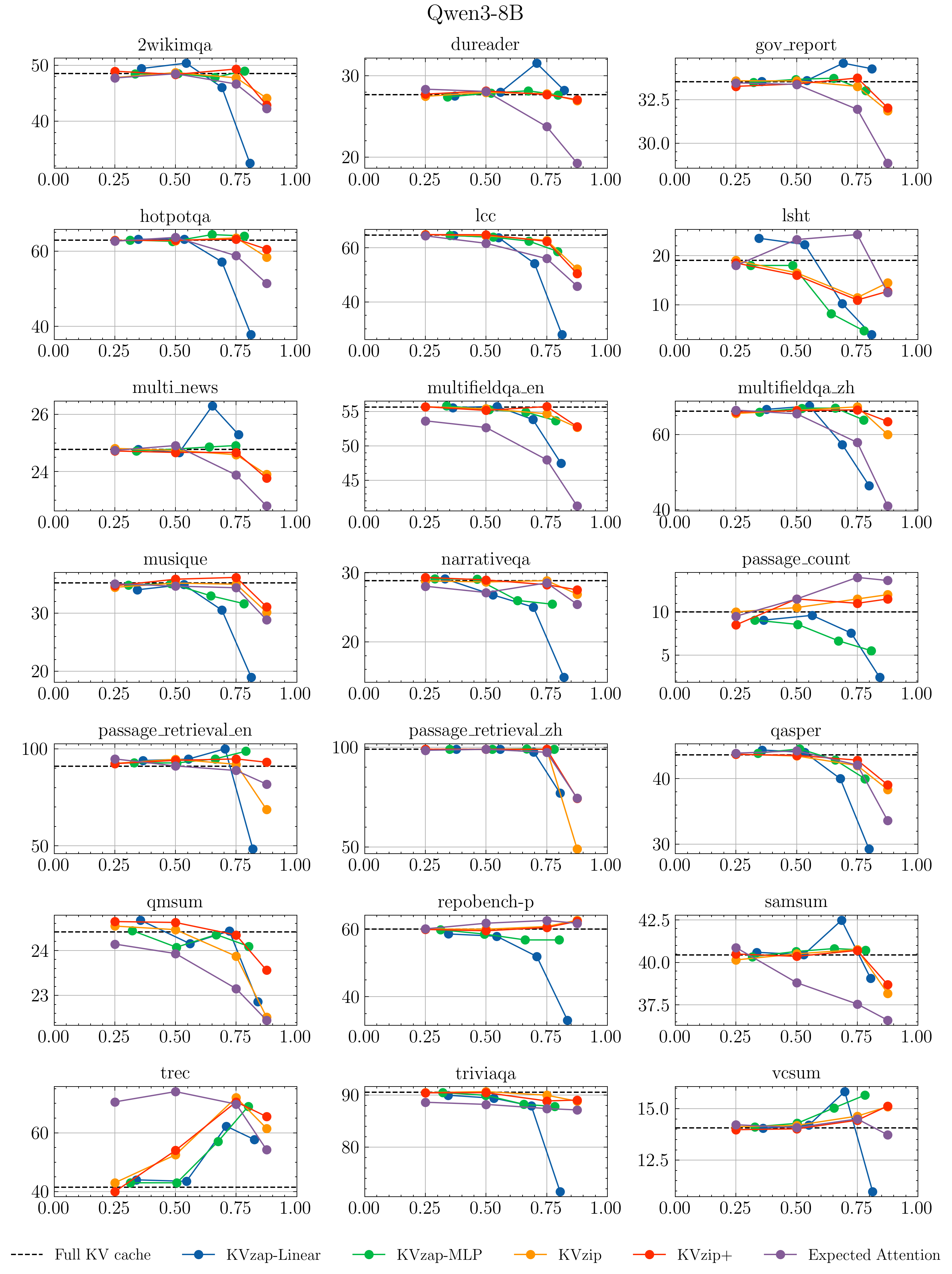}
    \caption{LongBench results for \qwen on each of the 21 subsets}
    \label{fig:longbench_details_qwen}
\end{figure}

\begin{figure}[t]
    \centering
    \includegraphics[width=\linewidth]{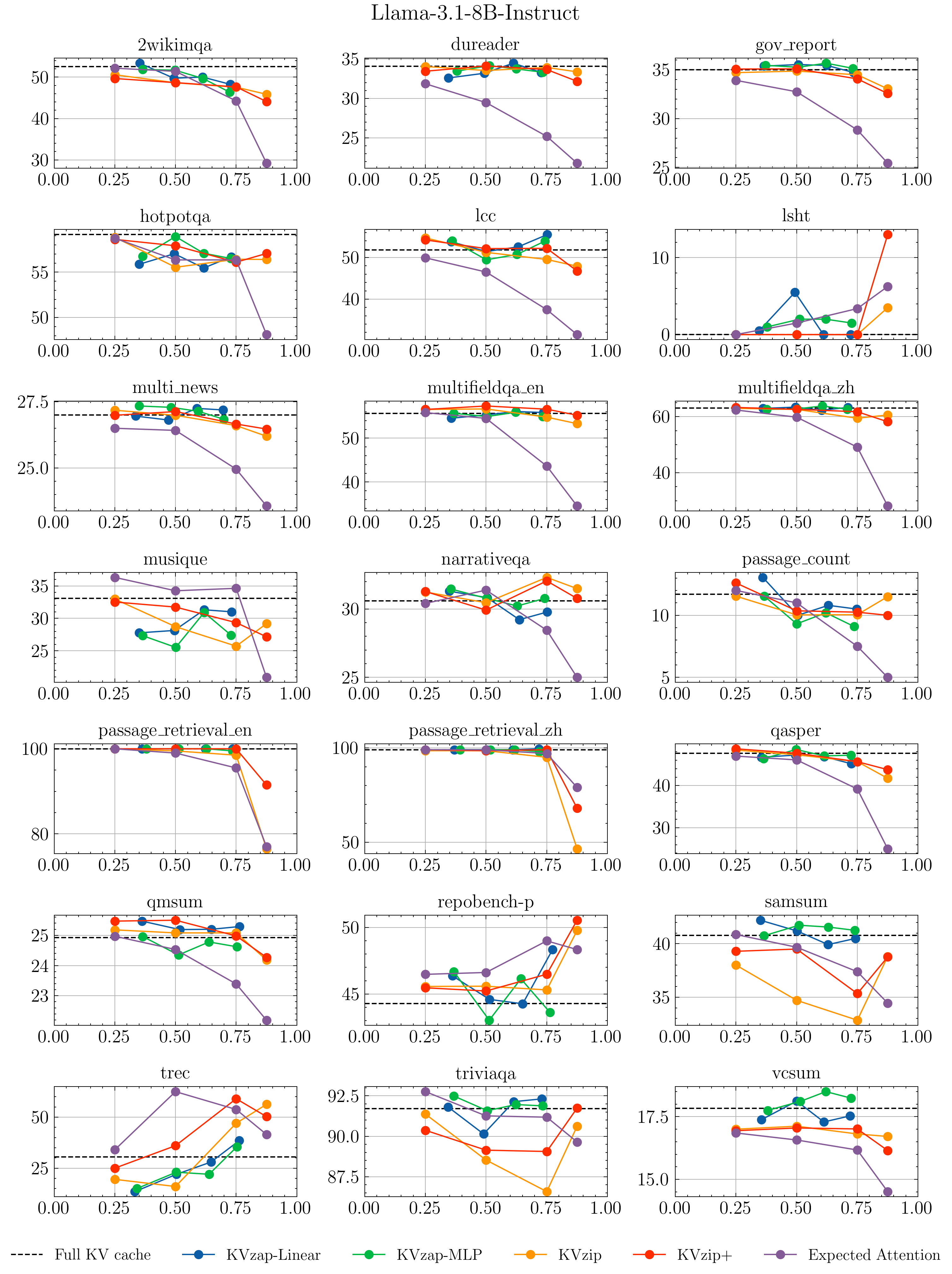}
    \caption{LongBench results for \llama on each of the 21 subsets}
    \label{fig:longbench_details_llama}
\end{figure}

\begin{figure}[t]
    \centering
    \includegraphics[width=\linewidth]{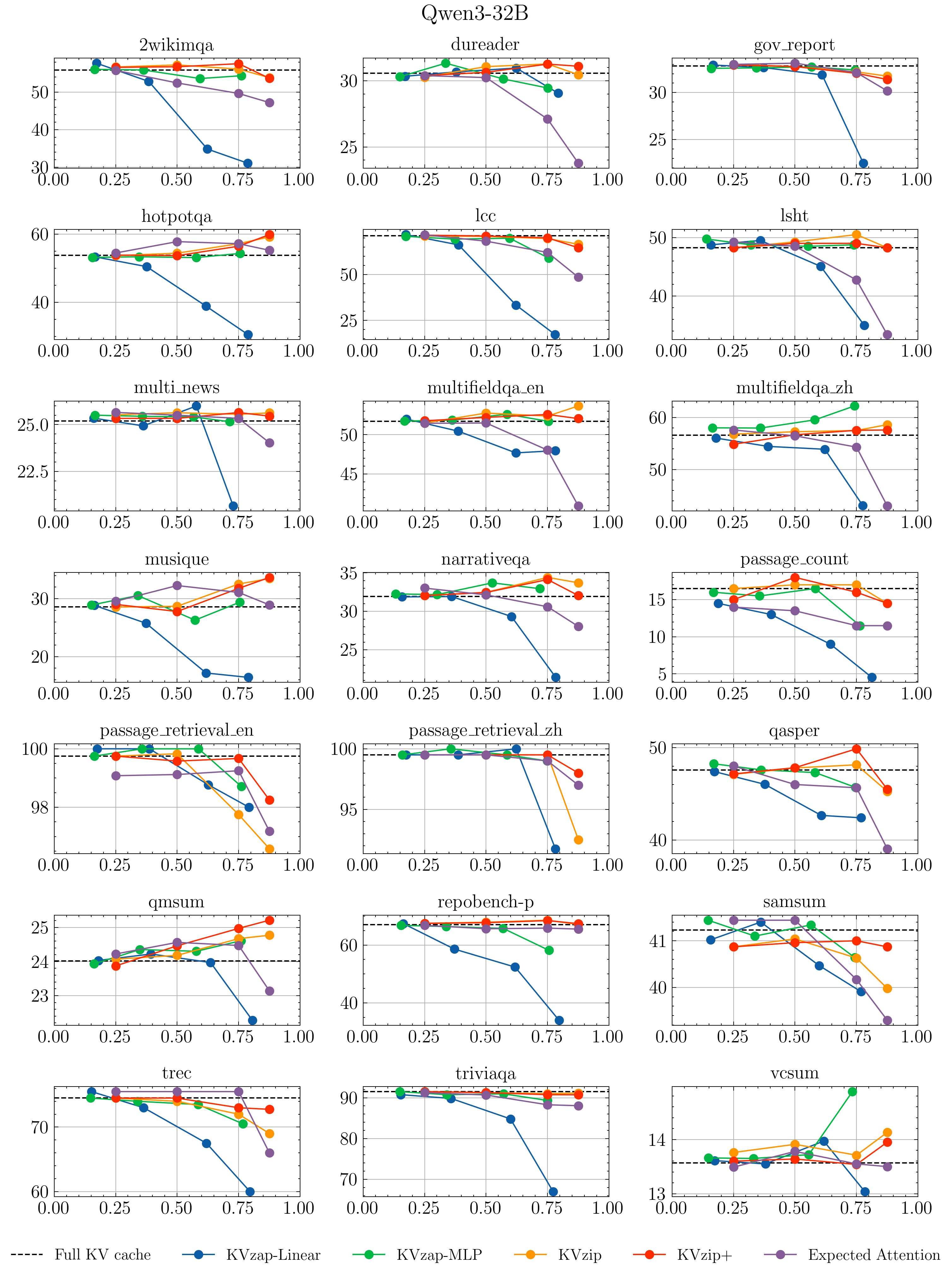}
    \caption{LongBench results for \qwenb on each of the 21 subsets}
    \label{fig:longbench_details_qwenb}
\end{figure}

Figures \ref{fig:longbench_details_qwen}--\ref{fig:longbench_details_qwenb} provide per-subset LongBench results (\qwen, \llama, \qwenb). We also report the average score excluding TREC in Figure \ref{fig:longbench_no_trec}.

\paragraph{AIME25} 

\begin{table}[t!]
\centering
\caption{AIME25 number of correct answers ($n=30$) for \qwen and \qwenb across four rollouts. KVzap-Linear with $\tau=-3$ achieved 96\% compression for \qwen and 93\% for \qwenb, explaining the zero scores.}
\label{tab:aime_results}
\begin{tabular}{lccc}
\toprule
\textbf{Method} & \textbf{Threshold ($\tau$)} & \textbf{Qwen3-8B} & \textbf{Qwen3-32B} \\
\midrule
\textbf{Full KV Cache} & - & 18, 20, 21, 21 & 19, 20, 22, 22 \\
\midrule
\multirow{4}{*}{\textbf{KVzap-Linear}} 
 & $-3$ & 0, 0, 0, 0 & 0, 0, 0, 0 \\
 & $-4$ & 10, 12, 13, 16 & 11, 11, 11, 13 \\
 & $-5$ & 17, 21, 21, 22 & 17, 21, 21, 22 \\
 & $-6$ & 19, 20, 21, 22 & 21, 22, 22, 23 \\
\midrule
\multirow{4}{*}{\textbf{KVzap-MLP}} 
 & $-3$ & 7, 9, 11, 11 & 16, 16, 17, 18 \\
 & $-4$ & 16, 17, 19, 20 & 20, 20, 22, 22 \\
 & $-5$ & 20, 20, 21, 22 & 20, 21, 22, 23 \\
 & $-6$ & 18, 20, 20, 21 & 22, 22, 23, 24 \\
\bottomrule
\end{tabular}
\end{table}

We report results for each of the four rollouts of AIME25 in Table \ref{tab:aime_results}.

\end{document}